%% file: egpaper_for_review.tex
\documentclass[10pt,twocolumn,letterpaper]{article}

\usepackage{iccv}
\usepackage{times}
\usepackage{epsfig}
\usepackage{graphicx}
\usepackage{amsmath}
\usepackage{amssymb}

\usepackage{multirow}
\usepackage[linesnumbered,ruled,vlined]{algorithm2e}
\SetKwInput{KwInput}{Input}                
\SetKwInput{KwOutput}{Output}              

\newcommand{\ProjectName}{DC3-GAN}

\usepackage[breaklinks=true,bookmarks=false]{hyperref}
\iccvfinalcopy 

\def\iccvPaperID{5490} 

\ificcvfinal\fi

\begin{document}

\title{Double cycle-consistent generative adversarial network for unsupervised conditional generation}

\author{Fei Ding, Feng Luo, Yin Yang\\
School of Computing, Clemson University\\
{\tt\small \{feid, luofeng, yin5\}@clemson.edu}
}

\maketitle
\ificcvfinal\thispagestyle{empty}\fi

\begin{abstract}

Conditional generative models have achieved considerable success in the past few years, but usually require a lot of labeled data. Recently, ClusterGAN combines GAN with an encoder to achieve remarkable clustering performance via unsupervised conditional generation. However, it ignores the real conditional distribution of data, which leads to generating less diverse samples for each class and makes the encoder only achieve sub-optimal clustering performance. Here, we propose a new unsupervised conditional generation framework, Double Cycle-Consistent Conditional GAN (\ProjectName), which can generate diverse class-conditioned samples. We enforce the encoder and the generator of GAN to form an encoder-generator pair in addition to the generator-encoder pair, which enables us to avoid the low-diversity generation and the triviality of latent features. We train the encoder-generator pair using real data, which can indirectly estimate the real conditional distribution. Meanwhile, this framework enforces the outputs of the encoder to match the inputs of GAN and the prior noise distribution, which disentangles latent space into two parts: one-hot discrete and continuous latent variables. The former can be directly expressed as clusters and the latter represents remaining unspecified factors. This work demonstrates that enhancing the diversity of unsupervised conditional generated samples can improve the clustering performance. Experiments on different benchmark datasets show that the proposed method outperforms existing generative model-based clustering methods, and also achieves the optimal disentanglement performance.

\end{abstract}

\section{Introduction}
\label{introduction}

Generative Adversarial Networks(GANs) have achieved remarkable success in realistic image generation such as class-conditioned generation~\cite{Karras2019stylegan2,Karras2020ada}, but at the cost of collecting massive amounts of annotated images. Given a large number of unlabeled images available online, how to leverage them for conditional generation remains a challenging problem. As an important unsupervised learning method, clustering has been widely used in many computer vision applications, such as image segmentation~\cite{chuang2006fuzzy}, visual features learning~\cite{caron2018deep}, and 3D object recognition~\cite{wang2019dominant}. Therefore, it's natural to combine clustering and generative models for an unsupervised conditional generation.


When processing high-semantic and high-dimensional images, most clustering methods such as DEC~\cite{xie2016unsupervised}, DCN~\cite{yang2017towards}, and ClusterGAN~\cite{mukherjee2019clustergan}, have been proposed to learn the `clustering-friendly' latent representations, then perform clustering algorithms, such as K-means~\cite{macqueen1967some} on the latent space. Since there are multiple optimization objectives, such as adversarial training for generation, low-dimensional representation learning, and distance-based clustering algorithms for class assignments, it's optimal to effectively integrate them to achieve unsupervised conditional generation in an end-to-end manner. Recently, the ClusterGAN~\cite{mukherjee2019clustergan} provides a new clustering mechanism based on the GAN and the inverse-mapping network. However, it only focuses on learning non-smooth latent space for clustering, and ignores the estimation of the real conditional distribution, leading to sub-optimal conditional generation.



In this paper, we introduce a novel unsupervised conditional generative model called Double Cycle-Consistent Conditional GAN (\ProjectName). It can generate diverse samples for each class without labels, and then directly obtains clusters without additional clustering methods. This framework can accommodate conditional generation and clustering in a unified and end-to-end manner. Moreover, we introduce a solution for directly obtaining class assignment by disentangling the latent space into two parts: the one-hot discrete latent variables directly related to categorical cluster information, and the continuous latent variables related to other factors of variations. The disentanglement of latent space is performing the clustering operation. Unlike the existing distance-based clustering methods, our method does not need any explicit clustering objectives or distance/similarity calculations in the latent space.

The conditional generation usually requires labels to estimate the real conditional distribution. But this work focuses on the unsupervised conditional generation, hence we propose to indirectly estimate the real conditional distribution via two cycle-consistencies: the generator-encoder pair and the encoder-generator pair. We first construct the generator-encoder pair with the generator of GAN and the encoder, which involves the mapping from latent space to data space, and back to latent space, to separate the latent space into one-hot discrete variables and continuous variables of other factors. Then, we utilize a weight sharing strategy to form a deterministic encoder-generator pair under the maximum mean discrepancy (MMD) regularization~\cite{gretton2012kernel}. Our method can be considered as the integration of the GAN and deterministic Autoencoder to achieve the unsupervised conditional generation. A better generator helps to guide the encoder for training, and a better encoder in turn helps to generate better class-conditioned samples. Therefore, we apply clustering as a proxy task to evaluate the estimation of the real conditional distribution. This framework includes three different types of regularizations: an adversarial density-ratio loss in data space, MMD loss in the continuous latent code, and cross-entropy loss in discrete latent code. The source code and models are publicly available at this link \footnote { after the paper is accepted}.


In summary, our contributions are as follows: 

(1) We propose a new unsupervised conditional GAN framework called \ProjectName, which can achieve effective conditional generation without labels, and directly obtain cluster assignments without clustering methods.

(2) We combine the encoder-generator pair with the generator-encoder pair to form two cycle-consistencies, which help avoid the triviality on continuous latent variable and enables estimation of the real conditional distribution. 

(3) We evaluate the conditional generation quality of \ProjectName, and apply it to different benchmark datasets for disentanglement and clustering. The experiments demonstrate that it can achieve desirable disentanglement and clustering performance in most cases.

\section{Method}
\label{method}

Given a collection i.i.d. samples $\mathbf{x} = \{x^{i}\}_{i=1}^N$ (\eg, images) drawn from an unknown data distribution $P_{\mathrm{x}}$, where $x^{i}$ is the $i$-th data sample and $N$ is the size of the dataset, the standard GAN~\cite{goodfellow2014generative,gulrajani2017improved} consists of two components: the generator $G_{\theta}$ and the discriminator $D_{\psi}$. $G_{\theta}$ defines a mapping from the latent space $\mathcal{Z}$ to the data space $\mathcal{X}$ and $D_{\psi}$ can be considered as a mapping from the data space $\mathcal{X}$ to the probability of one sample being real or not. To achieve unsupervised conditional generation, we need to introduce an inference network $E_{\phi}$ to obtain the latent variables given the data sample. 

In this section, we first conduct a comprehensive analysis of ClusterGAN~\cite{mukherjee2019clustergan}, and observe that there is a key loss item missing in the objective. To address this issue, we introduce an MMD-based regularization to enforce the inference network and the generator of standard GAN to form a deterministic Autoencoder. Meanwhile, the method enables us to disentangle the latent space $\mathbf{z}$ into the one-hot discrete latent variables $\mathbf{z}_c$, and the continuous latent variables $\mathbf{z}_n$ in an unsupervised manner. $\mathbf{z}_c$ naturally represents the categorical cluster information; $\mathbf{z}_n$ is expected to contain information of other variations.

\subsection{Unsupervised conditional generation}
\label{insight}

ClusterGAN~\cite{mukherjee2019clustergan} provides a new clustering method using GANs, which utilizes a joint distribution of discrete and continuous latent variables as the prior of GANs. Although it focuses on projecting the data to the latent space for clustering, it can be generalized to an unsupervised conditional generation framework. And the optimization is based on the combination of original GAN loss, cycle-consistency loss, and cross-entropy loss.


\begin{equation}
\begin{aligned}
{\min _{G,E} \max _{D} \mathcal{L}_{\operatorname{Clus}}(G,D,E) = }\\
{\underbrace{%
\mathbb{E}_{\mathbf{x} \sim P_{\mathrm{x}}}[q( D_{\psi}(\mathbf{x}))]+\mathbb{E}_{\mathbf{z}_c \sim P_{\mathrm{c}},\mathbf{z}_n \sim P_{\mathrm{n}}}[q(1-D_{\psi}(G_{\theta}(\mathbf{z}_c, \mathbf{z}_n)))]}_{\textcircled{\small{1}}}} \\
{- \lambda_n \underbrace{%
\mathbb{E}_{\mathbf{z}_c \sim P_{\mathrm{c}},\mathbf{z}_n \sim P_{\mathrm{n}}
}[c(E_{\phi}(G_{\theta}(\mathbf{z}_c, \mathbf{z}_n))_n,\mathbf{z}_n)]}_{\textcircled{\small{2}}}}\\
{- \lambda_c \underbrace{%
\mathbb{E}_{\mathbf{z}_c \sim P_{\mathrm{c}},\mathbf{z}_n \sim P_{\mathrm{n}}}[c(E_{\phi}(G_{\theta}(\mathbf{z}_c, \mathbf{z}_n))_c,\mathbf{z}_c)]}_{\textcircled{\small{3}}}
,}
\end{aligned}
\end{equation}

where $P_{\mathrm{x}}$ is the real data distribution, $P_{\mathrm{c}}$ is the prior distribution of $\mathbf{z}_c$, and $P_{\mathrm{n}}$ is the prior distribution of $\mathbf{z}_n$. $c(\cdot, \cdot)$ is any measurable cost function, $\lambda_n$ and $\lambda_c$ are hyperparameters balancing these losses. For the original GAN~\cite{goodfellow2014generative}, the function $q$ is chosen as $q(t)=\log t$, and the Wasserstein GAN~\cite{gulrajani2017improved} applies $q(t)=t$. This adversarial density-ratio estimation~\cite{tschannen2018recent} enforces $Q_{\mathrm{x}}$ to match $P_{\mathrm{x}}$, as shown in term $\textcircled{\small{1}}$, $\mathcal{L}_{\operatorname{GAN}}$. The term $\textcircled{\small{2}}$ and $\textcircled{\small{3}}$ are two constraints to the generator $G_{\theta}$ and the encoder $E_{\phi}$, which correspond to the cycle-consistency of $\mathbf{z}_n$ and the cross-entropy loss on $\mathbf{z}_c$.

To analyze this clearly, the term $\textcircled{\small{2}}$ can be written as:
\begin{equation}
\begin{aligned}
\mathcal{L}_n (G,E) = {- \mathbb{E}_{(\mathbf{x}, \mathbf{z}_n) \sim Q_{x c}}[c(E_{\phi}(\mathbf{x})_n,\mathbf{z}_n)] } \\
{ = \mathbb{E}_{\mathbf{z}_c \sim P_{\mathrm{c}},\mathbf{z}_n \sim P_{\mathrm{n}}
}[ || E_{\phi}(G_{\theta}(\mathbf{z}_c, \mathbf{z}_n)) - \mathbf{z}_n || ].}
\end{aligned}
\end{equation}
Thus, this loss term attempts to keep the cycle-consistency of $\mathbf{z}_n$
during optimization. After adding the recovery of $\mathbf{z}_n$, the information from $\mathbf{z}_n$ can be utilized for generation to a certain extent. However, since the dimension of $\mathbf{x}$ is much larger than the dimensions of $\mathbf{z}_c$ and $\mathbf{z}_n$, this constraint may become trivial for the generator-encoder (G-E) pair, and result in the generation of low-diversity samples.


The term $\textcircled{\small{3}}$ is  the cross-entropy loss on $\mathbf{z}_c$:
\begin{equation}\label{eqn:l_ce}
\mathcal{L}_{\operatorname{CE}} (G,E) = - \mathbb{E}_{(\mathbf{x}, \mathbf{z}_c) \sim Q_{x c}}[\log(Q^{E}(\mathbf{z}_c | \mathbf{x}))], 
\end{equation}
where $Q^{E}(\mathbf{z}_c | \mathbf{x})$ is used to denote the conditional distribution induced by $E_{\phi}$. $Q_{\mathbf{z}_c | \mathbf{x}}$ is the conditional distribution specified by the generator G. Therefore, minimizing loss term $\mathcal{L}_{\operatorname{CE}} (G,E)$ is equivalent to minimizing the KL divergence between $Q_{\mathbf{z}_c | \mathbf{x}}$ and $Q_{\mathbf{z}_c | \mathbf{x}}^{E}$. However, ClusterGAN ignores the real data conditional distributions $P_{\mathbf{z}_c | \mathbf{x}}$ in the objective, which usually requires real category information to estimate. Even when the marginal distributions $P_x$ and $Q_x$ match perfectly through the term $\textcircled{\small{1}}$, ClusterGAN still can not guarantee that two conditional distributions $P_{\mathbf{z}_c | \mathbf{x}}$ and $Q_{\mathbf{z}_c | \mathbf{x}}^{E}$ are well matched. Only minimizing $\mathcal{L}_{\operatorname{CE}} (G,E)$ makes G tend to generate data that are far from the decision boundaries of $E_{\phi}$. In other words, the generated images for each category may be easily distinguishable by $E_{\phi}$, but have low intra-class diversity. It is thus essential to incorporate $P_{\mathbf{z}_c | \mathbf{x}}$ in the objective function.

\begin{figure}
  \centering
  \includegraphics[width=0.47\textwidth]{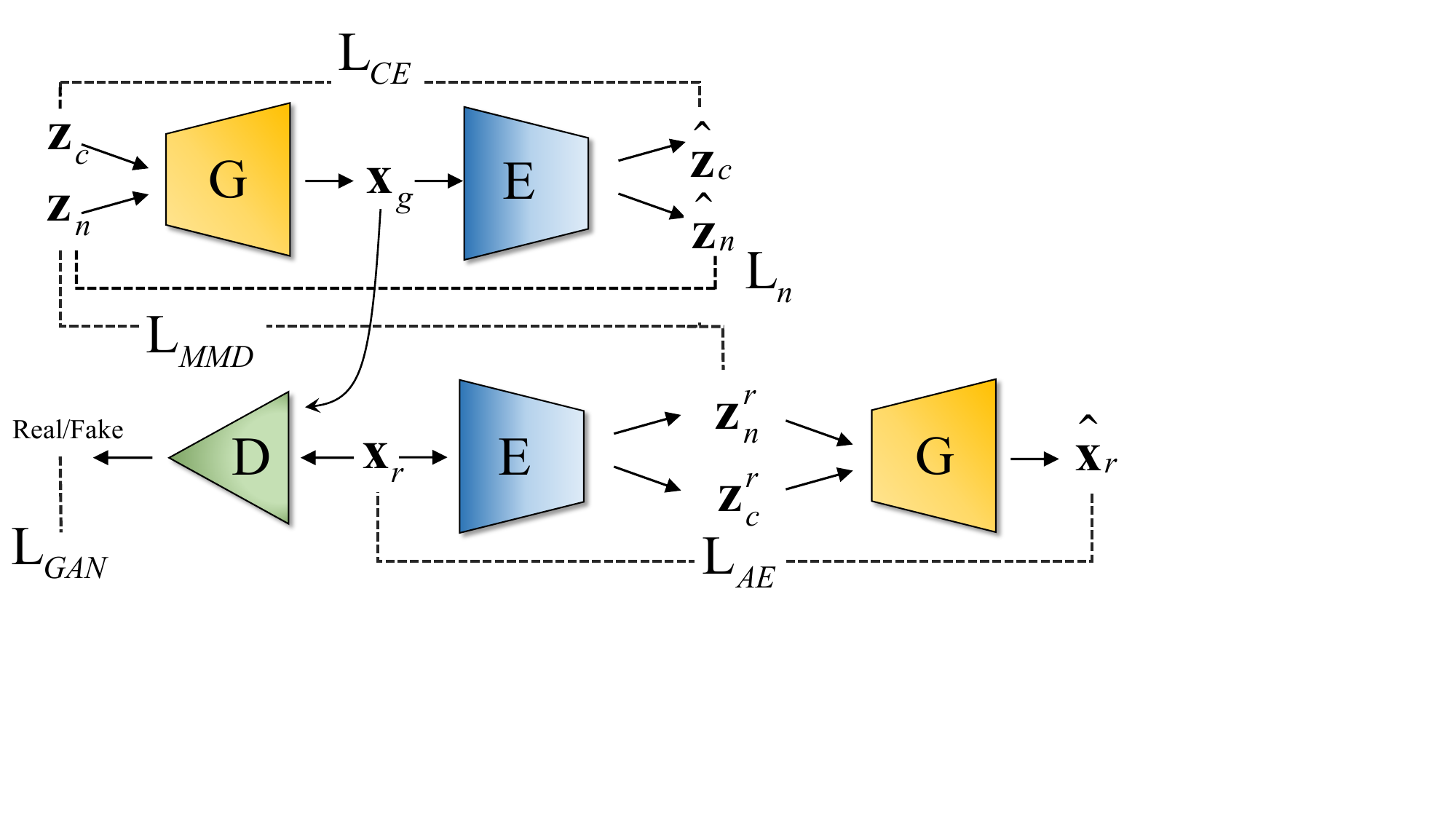}
  \caption{The architecture of~\ProjectName~(G: generator, E: encoder, D: discriminator). The latent representations are separated into one-hot discrete latent variables $\mathbf{z}_c$ and other factors of variation $\mathbf{z}_n$. The $\mathbf{z}_c$ and $\mathbf{z}_n$ are concatenated and fed into the $G_{\theta}$ for generation and the $E_{\phi}$ maps the samples ($\mathbf{x}_g$ and $\mathbf{x}_r$) back into latent space. The $D_{\psi}$ is adopted for the adversarial training in the data space. Note that all generators share the same parameters and all encoders share the same parameters. }
\label{fig:overview}  
\end{figure}

\subsection{The encoder-generator pair}\label{subsection:pair}


Our above analysis of ClusterGAN reveals that simply adding an encoder cannot effectively achieve conditional generation, which has two main problems: trivial continuous latent variables recovery and missing real conditional distribution term, $P_{\mathbf{z}_c | \mathbf{x}}$. Therefore, we present to enforce E and G to form an Autoencoder (E-G pair) by introducing a distance-based regularizer. The real conditional distribution $P_{\mathbf{z}_c | \mathbf{x}}$ can also be estimated properly in an unsupervised manner. We define the following objective:
\begin{equation}
\begin{aligned}
{\min _{G,E} \mathcal{L}_{\operatorname{E-G}}(G,E) = }\\ {\mathbb{E}_{Q_{\phi}(\mathbf{z}_n, \mathbf{z}_c | \mathbf{x})}\left[\log P_{\theta}(\mathbf{x} | \mathbf{z}_n, \mathbf{z}_c)\right] + \lambda \cdot \mathcal{D}_{z}\left(Q_{z}, P_{z}\right),}
\end{aligned}
\end{equation}
where $\lambda > 0$ is a hyperparameter, $\mathcal{D}_{z}$ is an arbitrary divergence between $Q_{z}$ and $P_{z}$, which encourages the encoded distribution $Q_{z}$ to match the prior $P_{z}$. Because the latent variables  $\mathbf{z}=(\mathbf{z}_c, \mathbf{z}_n)$, and the prior distribution $P_z(\mathbf{z}_c, \mathbf{z}_n) = P_c(\mathbf{z}_c)P_n(\mathbf{z}_n)$,  these constraints can be added by simply penalizing the discrete variables part and the continuous variables part separately. 

The constraint of continuous variables $\mathbf{z}_n$ can be considered to apply similar regularizations in the generative Autoencoder model like AAE~\cite{makhzani2015adversarial} and WAE~\cite{tolstikhin2017wasserstein}. The former uses the GAN-based density-ratio trick to estimate the KL-divergence of distributions~\cite{tschannen2018recent}, and the latter minimizes the distance between distributions based on Maximum Mean Discrepancy (MMD)~\cite{gretton2012kernel,li2015generative}. We choose adversarial density-ratio estimation for modeling the data space because it can handle complex distributions. MMD-based regularizer is stable for optimization and works well with multivariate normal distributions~\cite{tschannen2018recent}. Therefore, we choose MMD to quantify the distance between the prior distribution $P_{n}(\mathbf{z}_n)$ and the posterior distribution $Q_{n}(\mathbf{z}_n | \mathbf{x})$. Compared with WAE, we only penalize the continuous latent variables $\mathbf{z}_n$, not the whole latent variable. The regularizer $\mathcal{D}_{z}$ based on MMD is expressed as:

\begin{equation}
\begin{aligned}{
\mathcal{L}_{\operatorname{MMD}}(E)= \frac{1}{N(N-1)} \sum_{\ell \neq j} k\left(z_n^{\ell}, z_n^{j}\right)+
} \\ {
\frac{1}{N(N-1)} \sum_{\ell \neq j} k\left(\hat{z}_n^{\ell}, \hat{z}_n^{j}\right)-\frac{2}{N^{2}} \sum_{\ell, j} k\left(z_n^{\ell},  \hat{z}_n^{j}\right),
}\end{aligned}
\end{equation}

where $k(\cdot, \cdot)$ can be any positive definite kernel, $\{z_n^1, \dots, z_n^N\}$ are sampled from the prior distribution $P_{n}(\mathbf{z}_n)$,  $\hat{z}_n^{i}$ is sampled from the posterior distribution $Q_{n}(\mathbf{z}_n | \mathbf{x})$ and $x^i$ is sampled from the real data samples for $i=1, 2, \ldots, N$.

The constraint of $\mathbf{z}_c$ can't be applied explicitly without labels. Instead, we use a mean absolute error (MAE) criterion to estimate the encoding distribution $Q_{\phi}(\mathbf{z} | \mathbf{x})$ and the decoding distribution $P_{\theta}(\mathbf{x} | \mathbf{z})$, which are taken to be deterministic and can be replaced by $E_{\phi}$ and $G_{\theta}$, respectively.
\begin{equation}
    \mathcal{L}_{\operatorname{AE}}(E,G)= \mathbb{E}_{\mathbf{x} \sim P_{\mathrm{x}}} [ | \mathbf{x} - G_{\theta}(E_{\phi}(\mathbf{x}))| ].
\end{equation}


\subsection{The generator-encoder pair}\label{subsection:disentangle}

In addition to the encoder-generator pair, it also necessary to emphasize the generator-encoder pair for the disentanglement between discrete and continuous latent variables, as shown in Figure~\ref{fig:overview}. Most of the existing methods~\cite{hadad2018two, zheng2019disentangling, patacchiola2019autoencoders} leverage labels to achieve the disentanglement of various factors. This work attempts to encourage independence between $Q_{n}(\mathbf{z}_n | \mathbf{x})$ and $Q_{c}(\mathbf{z}_c | \mathbf{x})$ as much as possible without labels. 

We sample the latent variables $\mathbf{z} = (\mathbf{z}_c, \mathbf{z}_n)$ from the discrete-continuous prior, through the generator-encoder pair, it should output the identical discrete and continuous latent variables $(\hat{\mathbf{z}}_c, \hat{\mathbf{z}}_n)$. It enforces the generator to take advantage of extra information from $\mathbf{z}_c$. Besides, the recovery of latent variables ensure that outputs of the encoder $E_{\phi}$  are conditionally independent. When $E_{\phi}$ maps the real data sample $\mathbf{x}$ to latent representations $\mathbf{z}_c^r$ and $\mathbf{z}_n^r$, which are expected to be conditionally independent. The cross-entropy loss (Eq.~\ref{eqn:l_ce}) between $\mathbf{z}_c$ and $\hat{\mathbf{z}}_c$ can ensure that the latent variables $\hat{\mathbf{z}}_c$ only contain class-related information. Besides, to ensure that the latent variables $\hat{\mathbf{z}}_c$ or $\hat{\mathbf{z}}_c^r$ don't contain any class-related information, it is necessary to apply additional regularizers to penalize $\hat{\mathbf{z}}_n$ and $\hat{\mathbf{z}}_n^r$, which are related to the loss $\mathcal{L}_n$ and $\mathcal{L}_{\operatorname{MMD}}$. 




\subsection{Objective of \ProjectName}\label{subsection:objective}

The objective function of our approach is integrated into the following form: 
\begin{equation}\label{eqn:objective}
    \mathcal{L} = \mathcal{L}_{\operatorname{GAN}} + \mathcal{L}_{\operatorname{AE}} + \beta_1 \mathcal{L}_{\operatorname{MMD}} + \beta_2 \mathcal{L}_n + \beta_3 \mathcal{L}_{\operatorname{CE}}.
\end{equation}

where the regularization coefficients $\beta_1$ to $\beta_3 \ge 0$, balancing the weights of different loss terms. Each term of Eq.~\ref{eqn:objective} plays a different role for three components: generator $G_{\theta}$, discriminator $D_{\psi}$, and encoder $E_{\phi}$. Both $\mathcal{L}_{\operatorname{GAN}}$ and $\mathcal{L}_{\operatorname{AE}}$ are related to $G_{\theta}$ and $E_{\phi}$, which constrain the whole latent variables. The $\mathcal{L}_{\operatorname{GAN}}$ term is also related to $D_{\psi}$, which focuses on distinguishing the true data samples from the fake samples generated by $G_{\theta}$. $\mathcal{L}_{\operatorname{MMD}}$ and $ \mathcal{L}_n$ are related to continuous latent variables, and $\mathcal{L}_{\operatorname{CE}}$ and $\mathcal{L}_c$ are related to discrete latent variables. All these loss terms are used to ensure that our algorithm disentangles the latent space generated from encoder into cluster information and remaining unspecified factors. The training procedure of \ProjectName~applies jointly updating the parameters of $G_{\theta}$, $D_{\psi}$ and $E_{\phi}$, as described in Appendix.
We empirically set $\beta_1 = \beta_2$ to enable a reasonable adjustment of the relative importance of continuous and discrete parts.

\section{Experiments}

In this section, we perform a variety of experiments to evaluate the effectiveness of our proposed method, including clusters assignment via $\mathbf{z}_c$ and visualization studies of $\mathbf{z}_n$. We also conduct ablation experiments to understand the contribution of various loss terms.

\textbf{Data sets.} The clustering experiments are carried out on seven datasets: MNIST~\cite{lecun1998gradient}, Fashion-MNIST~\cite{xiao2017fashion}, YouTube-Face (YTF)~\cite{wolf2011face}, Pendigits~\cite{alimoglu1996methods}, 10x\_73k~\cite{zheng2017massively}, COIL-100~\cite{nene1996columbia}, and  CIFAR-10~\cite{krizhevsky2009learning}. The disentanglement evaluation is conducted on dsprites~\cite{matthey2017dsprites}. Both of the first two datasets contain 70k images with 10 categories, and each sample is a $28 \times 28$ grayscale image. YTF contains 10k face images of size $55 \times 55$, belonging to 41 categories. The Pendigits dataset contains a time series of $(x, y)$ coordinates of hand-written digits. It has 10 categories and contains 10992 samples, and each sample is represented as a 16-dimensional vector. The 10x\_73k dataset contains 73233 data samples of single-cell RNA-seq counts of 8 cell types, and the dimension of each sample is 720. The multi-view object image dataset COIL-100 has 100 clusters and contains 7200 images of size $128 \times 128$.


\textbf{Implementation Details.} We implement different neural network structures for $G_{\theta}$, $D_{\psi}$, and $E_{\phi}$ to handle different types of data. We provide details of models in the Appendix. For the prior distribution of our method, we randomly generate the discrete latent code $\mathbf{z}_c$, which is equal to one of the elementary one-hot encoded vectors in $\mathbb{R}^K$, then we sample the continuous latent code from $\mathbf{z}_n \sim \mathcal{N}(\mathbf{0}, \mathbf{\sigma}^2 \mathbf{I}_{d_n})$, here $\sigma = 0.10$. The sampled latent code $\mathbf{z}=(\mathbf{z}_c, \mathbf{z}_n)$ is used as the input of $G_{\theta}$ to generate samples. The dimensions of $\mathbf{z}_c$ and $\mathbf{z}_n$ are shown in Table~\ref{table:dimension}. We implement the MMD loss with RBF kernel~\cite{tolstikhin2017wasserstein} to penalize the posterior distribution $Q_{\phi} (\mathbf{z}_n | \mathbf{x})$. The improved GAN variant with a gradient penalty~\cite{gulrajani2017improved} is used in all experiments. To obtain the cluster assignment, we directly use the argmax over all softmax probabilities for different clusters. The following regularization parameters work well during all experiments: $\lambda=10$, $\beta_1=\beta_2=0.1$, $\beta_3=10$. We implement the models using the TensorFlow library and train them on one NVIDIA DGX-1 station.

\subsection{Evaluation of generation quality}

\begin{table}[!ht]
\centering
\caption{Comparison of FID score to reveal the quality of generated samples from GAN methods (Lower is better).}
\label{table:fid}
\begin{tabular}{ccccc}\\
\hline  
Method & Ours & ClusterGAN & WGAN & InfoGAN\\
\hline
MNIST & \textbf{0.15} & 0.81 & 0.88 & 1.88\\  
\hline
Fashion & \textbf{0.67} & 0.91 & 0.95 & 11.04\\ 
\hline
\end{tabular}
\end{table}

\begin{table*}[!ht]
\centering
\caption{Comparison of mean SSIM scores of 200 pairs to reveal the diversity of generated samples from GAN methods (Lower is better).}
\label{table:ssim}
\begin{tabular}{ccccccccccc}\\
\hline  
Class & 0 & 1 & 2 & 3 & 4 & 5 & 6 & 7 & 8 & 9\\
\hline
ClusterGAN & 0.362 & 0.599 & 0.263 & \textbf{0.314} & 0.315 & 0.282 & 0.351 & \textbf{0.388} & 0.340 & 0.427 \\ 
\hline
Ours & \textbf{0.343} & \textbf{0.576} & \textbf{0.231} & 0.316 & \textbf{0.312} & \textbf{0.259} & \textbf{0.322} & 0.392 & \textbf{0.336} & \textbf{0.377} \\ 
\hline
\end{tabular}
\end{table*}

%

\begin{figure}[t]
\begin{center}
 \includegraphics[width=\linewidth]{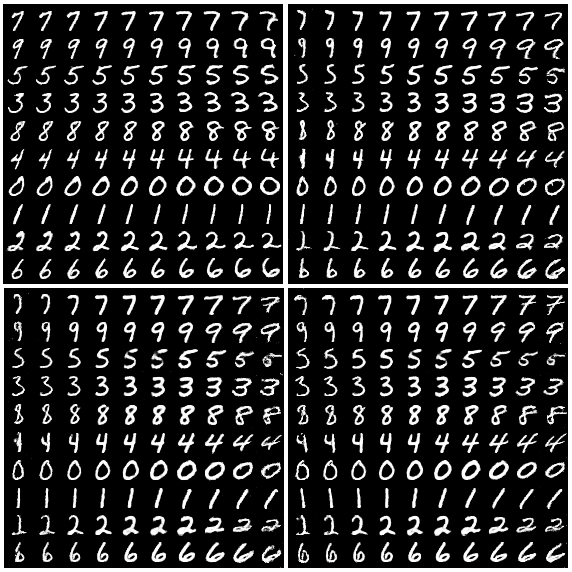}
\end{center}
   \caption{ Samples generated on fixed discrete latent codes from the models trained on MNIST. }
\label{fig:traversal}
\end{figure}

To demonstrate the quality and diversity of generated samples from~\ProjectName, we first calculate the Frechet Inception Distance (FID)~\cite{heusel2017gans} score of generated samples, as shown in Table~\ref{table:fid}. The FID scores on MINST and Fashion are significantly lower than those of ClusterGAN. Our method shows that the estimation of real conditional distribution can improve the quality of generated samples. Then we randomly sample 200 pairs of generated images from one category to calculate structural similarity (SSIM)~\cite{wang2004image,wang2009mean} for diversity evaluation on MNIST data. This evaluation method for diversity has also been used in AC-GAN~\cite{odena2017conditional}. The SSIM scores range between 0.0 and 1.0, and lower mean scores indicate that samples from the same class are less similar. As shown in Table~\ref{table:ssim}, our method achieves lower SSIM scores on most classes, which demonstrates that it can enhance the diversity of generation. The diversity of generated images indicates that there exist different latent variables for generative factors, except the cluster information. To further understand these generative factors, we change the value of one single dimension from  $[-0.5, 0.5]$ in $\mathbf{z}_n$ while fixing other dimensions and the discrete latent variables $\mathbf{z}_c$. As shown in Figure~\ref{fig:traversal}, the value changing leads to semantic changes in generated samples. The changed dimensions represent the tilt, style, and width factors of digits, which shows the potential to disentangle the latent space.

\subsection{Evaluation of disentanglement}

We further explore the disentanglement capability of~\ProjectName on dSprites dataset. We follow the same experimental settings and hyperparameters tuning as FactorVAE~\cite{kim2018disentangling}, InfoGAN~\cite{chen2016infogan} and InfoGAN-CR~\cite{lin2020infogan} for fair comparisons. We provide the experimental details in Appendix, and focus on explaining the results in this section. As shown in Table~\ref{table:disentanglement}, our method also achieves excellent disentanglement performance. Compared with InfoGAN-CR, we implement the proposed double-cycle consistency to replace the contrastive regularizer (CR) based on the InfoGAN architecture, which has two latent variables. These consistencies force the generator to generate different samples while fixing one latent variable and changing another latent variable. This is beneficial for disentanglement, as it simulates the latent traversal experiments and encourages distinct changes in generated samples. In addition, ModelCentrality is proposed by~\cite{lin2020infogan} for unsupervised model selection to evaluate the trained models on an unlabelled dataset. It's naturally suitable for our unsupervised conditional generation settings.

\begin{table*}[!ht]
\centering
\caption{Comparison results based on different disentanglement metrics on the dSprites dataset.The score 1.0 denotes a perfect disentanglement. All the baseline results are from~\cite{lin2020infogan}. The proposed~\ProjectName~achieves desirable scores in most cases. The implementation of~\ProjectName~is based on the source code of InfoGAN-CR, and MC (ModelCentrality) denotes an unsupervised model selection scheme~\cite{lin2020infogan}.}
\label{table:disentanglement}
\begin{tabular}{lccccccc}
\hline
Model & FactorVAE & DCI & SAP & Explicitness & Modularity & MIG & BetaVAE\\
\hline
VAE & 0.63 $\pm$ 0.06 & 0.30 $\pm$ 0.10 & - & - & - & 0.10 & - \\
$\beta$-TCVAE & 0.62 $\pm$ 0.07 & 0.29 $\pm$ 0.10 & - & - & - & \textbf{0.45} & - \\
HFVAE & 0.63 $\pm$ 0.08 & 0.39 $\pm$ 0.16 & - & - & - & - & - \\
$\beta$-VAE & 0.63 $\pm$ 0.10 & 0.41 $\pm$ 0.11 & 0.55 & - & - & 0.21 & - \\
FactorVAE  & 0.82 & - & - & - & - & 0.15 & - \\
FactorVAE (1.0) & 0.79 $\pm$ 0.01 & 0.67 $\pm$ 0.03 & 0.47 $\pm$ 0.03 & 0.78 $\pm$ 0.01 & 0.79 $\pm$ 0.01 & 0.27 $\pm$ 0.03 & 0.79 $\pm$ 0.02 \\
FactorVAE (10.0) & 0.83 $\pm$ 0.01 & 0.70 $\pm$ 0.02 & 0.57 $\pm$ 0.0 & 0.79 $\pm$ 0.0 & 0.79 $\pm$ 0.0 & 0.40 $\pm$ 0.01 & 0.83 $\pm$ 0.0 \\
FactorVAE (40.0) & 0.82 $\pm$ 0.01 & 0.74 $\pm$ 0.01 & 0.56 $\pm$ 0.0 & 0.79 $\pm$ 0.0 & 0.77 $\pm$ 0.01 & 0.43 $\pm$ 0.01 & 0.84 $\pm$ 0.01 \\
FactorVAE + MC & 0.84 $\pm$ 0.0 & 0.73 $\pm$ 0.01 & 0.58 $\pm$ 0.0 & 0.80 $\pm$ 0.0 & 0.82 $\pm$ 0.0 & 0.37 $\pm$ 0.0 & 0.86 $\pm$ 0.0 \\
IB-GAN  & 0.80 $\pm$ 0.07 & 0.67 $\pm$ 0.07 & - & - & - & - & - \\
InfoGAN  & 0.82 $\pm$ 0.01 & 0.60 $\pm$ 0.02 & 0.41 $\pm$ 0.02 & 0.82 $\pm$ 0.0 & 0.94 $\pm$ 0.01 & 0.22 $\pm$ 0.01 & 0.87 $\pm$ 0.01 \\
InfoGAN-CR + MC  & 0.92 $\pm$ 0.0 & 0.77 $\pm$ 0.0 & \textbf{0.65 $\pm$ 0.0} & \textbf{0.87 $\pm$ 0.0} & \textbf{0.99 $\pm$ 0.0} & \textbf{0.45 $\pm$ 0.0} & 0.99 $\pm$ 0.0 \\
Ours + MC  & \textbf{0.936 $\pm$ 0.0} & \textbf{0.790 $\pm$ 0.0} & 0.634 $\pm$ 0.0 & 0.862 $\pm$ 0.0 & 0.985 $\pm$ 0.0 & 0.378 $\pm$ 0.0 & \textbf{0.998 $\pm$ 0.0} \\
\hline
\end{tabular}
\end{table*}

\subsection{Evaluation of clustering algorithm}

\begin{table*}
  \caption{Comparison of clustering algorithms on five benchmark datasets. The results marked by (*) are from existing sklearn.cluster.KMeans package. The dash marks (-) mean that the source code is not available or that running released code is not practical, all other results are from \cite{mukherjee2019clustergan} and \cite{yang2019deep}. SpecNet and ClusGAN mean SpectralNet and ClusterGAN.}
  \label{table:clustering}
  \centering
\begin{tabular}{@{}lcccccccccc@{}}
\hline
\multirow{2}{*}{Method} & \multicolumn{2}{c}{MNIST} & \multicolumn{2}{c}{Fashion-10} &
          \multicolumn{2}{c}{YTF} & \multicolumn{2}{c}{Pendigits} & \multicolumn{2}{c}{10x\_73k} \\
\cline{2-11}
 & ACC & NMI & ACC & NMI & ACC & NMI & ACC & NMI & ACC & NMI\\
\hline
K-means & 0.532 & 0.500 & 0.474 & 0.512 & 0.601 & 0.776 & 0.793$^{*}$ & 0.730$^{*}$ & 0.623$^{*}$ & 0.577$^{*}$ \\
NMF & 0.560 & 0.450 & 0.500 & 0.510 & - & - & 0.670 & 0.580 & 0.710 & 0.690 \\
DEC& 0.863 & 0.834 & 0.518 & 0.546 & 0.371 & 0.446 & - & - & - & - \\
DCN& 0.830 & 0.810 & - & - & - & - & 0.720 & 0.690 & - & - \\
JULE& 0.964 & 0.913 & 0.563 & 0.608 & 0.684 & 0.848 & - & - & - & - \\
DEPICT& 0.965 & 0.917 & 0.392 & 0.392 & 0.621 & 0.802 & - & - & - & - \\
SpecNet& 0.800 & 0.814 & - & - & 0.685 & 0.798 & - & - & - & - \\
InfoGAN& 0.890 & 0.860 & 0.610 & 0.590 & - & - & 0.720 & 0.730 & 0.620 & 0.580 \\
ClusGAN& 0.950 & 0.890 & 0.630 & 0.640 & - & - & 0.770 & 0.730 & 0.810 & 0.730 \\
DualAE& \textbf{0.978} & \textbf{0.941} & 0.662 & 0.645 & 0.691 & \textbf{0.857} & - & - & - & - \\
Ours & 0.976 & 0.941 & \textbf{0.693} & \textbf{0.669} & 0.\textbf{721} & 0.790 & \textbf{0.847} & \textbf{0.803} & \textbf{0.905} & \textbf{0.820}\\
\hline

\end{tabular}
\end{table*}

We argue that the clustering task can be considered as a proxy to evaluate the real conditional distribution estimation. To evaluate clustering results, we report two standard evaluation metrics: Clustering Purity (ACC) and Normalized Mutual Information (NMI). We compare \ProjectName \, with four clustering baselines: K-means~\cite{macqueen1967some}, Non-negative Matrix Factorization (NMF)~\cite{lee1999learning}.
We also compare our method with the state-of-the-art clustering approaches based on GAN and Autoencoder, respectively. For GAN-based approaches, ClusterGAN~\cite{mukherjee2019clustergan} is chosen as it achieves the superior clustering performance compared to other GAN models (\eg, InfoGAN). For Autoencoder-based methods such as DEC~\cite{xie2016unsupervised}, DCN~\cite{yang2017towards} and DEPICT~\cite{ghasedi2017deep}, Dual Autoencoder Network (DualAE)~\cite{yang2019deep} are used for comparison. In addition, the deep spectral clustering (SpectralNet)~\cite{shaham2018spectralnet} and joint unsupervised learning (JULE)~\cite{yang2016joint} are also included in the comparison.

Table~\ref{table:clustering} reports the best clustering metrics of different models from 5 runs. Our method achieves significant performance improvement on Fashion-10, YTF, Pendigits, and 10x\_73k datasets than other methods. Particularly, while all other methods perform worse than K-means on the 16-dimensional Pendigit dataset, our method significantly outperforms K-means in both ACC (0.847 vs. 0.793) and NMI (0.803 vs. 0.730). \ProjectName~ achieves the best ACC result on YTF dataset while maintaining comparable NMI value. For MNIST dataset, \ProjectName~achieves close to the best performance on both ACC and NMI metrics. To further evaluate the performance of \ProjectName~ on large numbers of clusters, we compare our clustering method with K-means on Coil-100 dataset using three standard evaluation metrics: ACC, NMI, and Adjusted Rand Index (ARI). As shown in Table~\ref{table:scalability}, \ProjectName~ achieves better performance on all three metrics.


\subsection{Evaluation on more images}

\begin{figure*}
  \centering
  \includegraphics[width=0.85\textwidth]{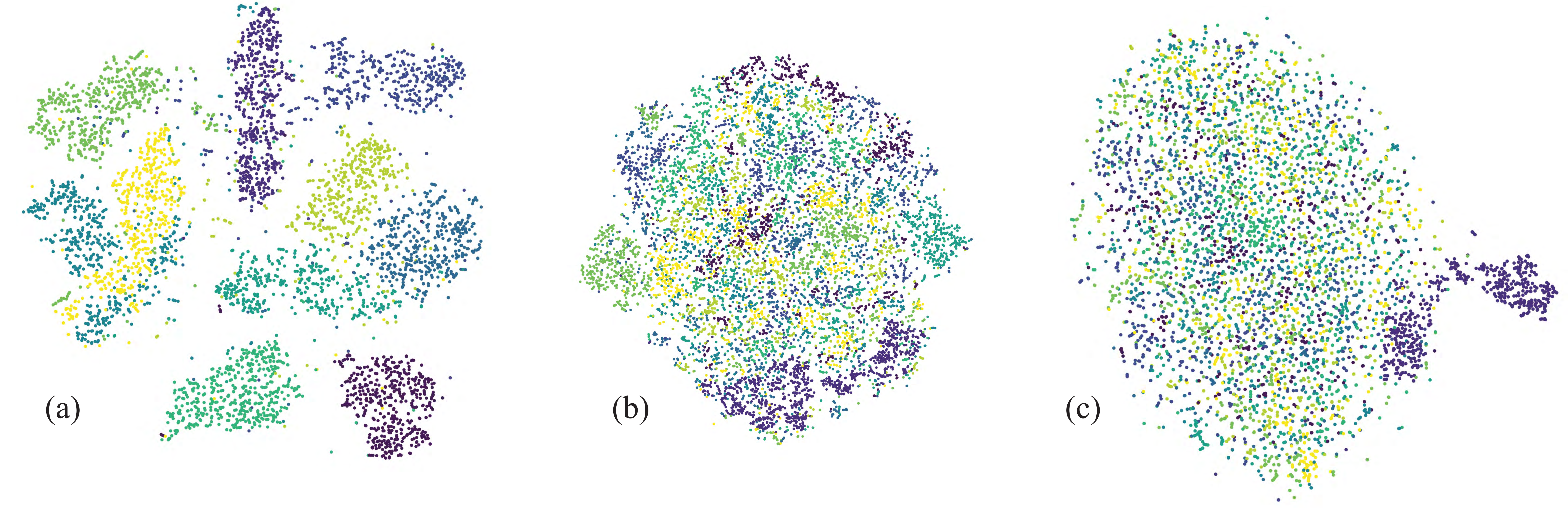}
  \caption{The t-SNE visualization of raw data (a),  $\mathbf{z}_n$ of ClusterGAN (b) and \ProjectName~(c) on MNIST dataset. The bulk of samples in the right part of a(3) is a small group of ``1'' images. The reason that they are not well mixed may be due to their low complexity.}
\label{fig:tsne}  
\end{figure*}

We also use the t-SNE~\cite{maaten2008visualizing} algorithm to visualize $\mathbf{z}_n$ of MNIST datasets and compare them to ClusterGAN and the original data. As shown in Figure~\ref{fig:tsne}, we can observe different categories in the original data. In ClusterGAN, there are still several distinguishable clusters. In contrast, our method can make these points more cluttered in latent space, which doesn't contain obvious category information in the $\mathbf{z}_n$. Therefore, our method demonstrates another excellent capability: all these informative continuous factors are independent of cluster information.

\begin{table}[!ht]
\centering
\caption{The clustering results on the Coil-100 dataset, which has a large number of clusters (K=100).}
\label{table:scalability}
\begin{tabular}{lccc }
\hline
Method & ACC & NMI & ARI \\
\hline
K-means & 0.668 & 0.836 & 0.574 \\
ClusterGAN & 0.615 & 0.797 & 0.487 \\
Our method & \textbf{0.822} & \textbf{0.911} & \textbf{0.764} \\
\hline
\end{tabular}
\end{table}

\begin{figure}[t]
\begin{center}
\includegraphics[width=\linewidth]{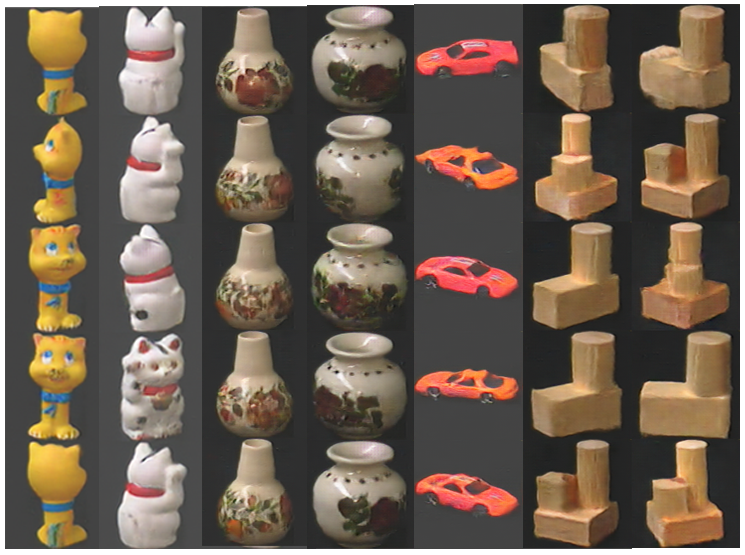}
\end{center}
   \caption{The samples generated on fixed discrete latent variables from the models trained on Coil-100 dataset. Each column corresponds to a specific cluster.}
\label{fig:scalability}
\end{figure}

We first evaluate the scalability of \ProjectName~ to large numbers of clusters on the COIL-100 dataset(100 clusters). Here, we compare our clustering method with K-means on three standard evaluation metrics: ACC, NMI and Adjusted Rand Index (ARI). As shown in Table~\ref{table:scalability}, \ProjectName~ achieves better performance on all three metrics.  \ProjectName~ even gains an increase of 0.154 on ACC metric. We also perform image generation task on Coil-100 dataset, to further verify the generative performance, which involves mapping latent variables to the data space. Figure~\ref{fig:scalability} shows the generated samples by fixing one-hot discrete latent variables, which are diverse and realistic. The continuous latent variables represent meaningful factors such as the pose, location and orientation information of objects. Therefore, the disentanglement of latent space not only provides the superior clustering performance, but also retains the remarkable ability of diverse and high-quality image generation. 

Besides, we further evaluate the proposed method on more complex dataset: CIFAR-10. The implementation is based on Google compare-gan framework~\footnote{\url{https://github.com/google/compare_gan}}. 
The spectral normalization is used on both generator and discriminator. We use the same class-conditional BatchNorm in the generator as Lucic~\etal~\cite{lucic2019high}, to incorporate the category information from $\mathbf{z}_n$. For the encoder, we use the pre-trained SimCLR~\cite{chen2020simple} model to improving training efficiency, and apply 2-layer MLP as project head to map the learned representations to $\mathbf{z}_n$ and $\mathbf{z}_c$. The self-supervised SimCLR model is pre-trained by following the official implementation~\footnote{\url{https://github.com/google-research/simclr}}. Table~\ref{table:more-clustering} shows that~\ProjectName~achieves close to the best clustering performance on ACC. Because our method learns cluster memberships from  conditional generation without labels, it's also necessary to evaluate the generation results of images. As shown in Table~\ref{table:fid-score}, our method also maintains the quality of image generation, which enables to achieve the superior clustering results.

\begin{table}[th]
\caption{CIFAR-10 images clustering results. All baseline results are from \cite{ji2019invariant}. The value marked by (*) is the best (mean) results in~\cite{ji2019invariant}, and they also report that avg. $\pm$ STD is 0.576 $\pm$ 0.050.}
\label{table:more-clustering}
\centering
\begin{tabular}{lcc}
\hline
Method & ACC & NMI\\
\hline
K-means & 0.229 & 0.087 \\
DCGAN (2015)~\cite{radford2015unsupervised} & 0.315 & 0.265 \\
JULE (2016)~\cite{yang2016joint} & 0.272 & 0.192\\
DEC (2016)~\cite{xie2016unsupervised} & 0.301 & 0.257\\
DAC (2017)~\cite{chang2017deep} & 0.522 & 0.396\\
DeepCluster (2018)~\cite{caron2018deep} & 0.374 & -\\
ADC (2018)~\cite{haeusser2018associative} & 0.325 & -\\
IIC (2019)~\cite{ji2019invariant} & 0.617 (0.576)$^{*}$ & 0.513\\
GATCluster(2020)~\cite{niu2020gatcluster} & 0.610 & 0.475 \\
Ours &  0.605 & 0.484\\
\hline
\end{tabular}
\end{table}

\begin{table}
  \caption{FID results on the CIFAR-10 dataset (smaller is better). The results marked by (*) are from \cite{mao2019mode}.}
  \label{table:fid-score}
  \centering
  \begin{tabular}{lc}
    \hline
    Method     & FID Score     \\
    \hline
    DCGANs~\cite{radford2015unsupervised}   & 29.7$^{*}$ \\
    WGAN-GP (2017)~\cite{gulrajani2017improved}    & 29.3  \\
    SN-SMMDGAN (2018)~\cite{arbel2018gradient} & 25.0  \\
    MSGAN (2019)~\cite{mao2019mode}	  & 28.7$^{*}$  \\
    Ours              & 28.5 $\pm$ 0.02\\
    \hline
  \end{tabular}
\end{table}

\subsection{Ablative Analysis}

We perform the ablative analysis of our losses (Table~\ref{table:ablation}). The $\mathcal{L}_{\operatorname{AE}}$ and $\mathcal{L}_{\operatorname{MMD}}$ are critical in our model. The inference network and the generator form a deterministic encoder-decoder pair. To minimize the reconstruction loss $\mathcal{L}_{\operatorname{AE}}$, the generator $G_{\theta}$ needs to learn to generate realistic and diverse data samples. It also indirectly forces the $\mathbf{z}_c^r$ to contain only the category information. $\mathcal{L}_{\operatorname{MMD}}$ enforces the posterior distribution $Q_{\phi}(\mathbf{z}_n | \mathbf{x})$ to be close to the prior distribution $P(\mathbf{z}_n)$. The clustering performance gain is also from the loss terms $\mathcal{L}_{\operatorname{CE}}$ and $\mathcal{L}_{\operatorname{n}}$.

\begin{table}[]
\centering
\caption{Ablations on MNIST dataset. Each row shows the removal of a loss term. The full setting includes all loss terms.}
\label{table:ablation}
\begin{tabular}{lcc}
\hline
Ablative analysis  &  ACC & NMI\\
\hline
No $\mathcal{L}_{\operatorname{CE}}$ & 0.899 & 0.863 \\
No  $\mathcal{L}_n$ & 0.868 & 0.0.851 \\
No $\mathcal{L}_{\operatorname{MMD}}$ & 0.812 & 0.829 \\
No  $\mathcal{L}_{\operatorname{AE}}$ & 0.672 & 0.488 \\
Full setting & \textbf{0.976} & \textbf{0.941} \\
\hline
\end{tabular}
\end{table}

\section{Related works}

\textbf{Latent space clustering.} 
A general method to avoid the curse of dimensionality in clustering is mapping data samples to in a low-dimensional latent space and performing clustering on latent space. 
Most existing latent space clustering methods are based on Autoencoder~\cite{xie2016unsupervised, dilokthanakul2016deep, guo2017improved, yang2017towards,yang2019deep}, which enables reconstructing data samples from the low-dimensional representation. The training objectives usually are coupled with reconstruction loss to avoid random discriminative representations. However, it forces latent representations to capture all key factors of variations and similarities related to reconstruction, not class decision boundaries. Alternatively, recent contrastive learning utilizes instance discrimination to achieve remarkable self-supervised representation learning~\cite{chen2020simple}. But it still has potential limitations that instance-level pseudo labels are from hand-crafted augmentations and cannot explicitly determine the underlying class boundaries. Such strategies may benefit from the good parameter initialization, but there is still a lack of stable supervision signals to directly improve class assignments. Furthermore, these latent space clustering methods still depend on additional distance-based clustering algorithms (\eg, K-means) to obtain the cluster assignments. In contrast, our method accommodates the conditional generation and cluster assignments in an end-to-end manner.




\textbf{Generative models.} 
Learning disentangled representation can reveal the factors of variation in the data~\cite{bengio2013representation}, and provides interpretable semantic latent codes for generative models. Generally, existing models can be mainly categorized into VAE-based and GAN-based types. The VAE-based methods involves extracting the label relevant and irrelevant representations~\cite{mathieu2016disentangling, hadad2018two, zheng2019disentangling, patacchiola2019autoencoders}. For example,  Mathieu \etal~\cite{mathieu2016disentangling} introduce a conditional VAE with adversarial training to disentangle the latent representations into label relevant and the remaining unspecified factors. Y-AE~\cite{patacchiola2019autoencoders} focuses on the standard Autoencoder to achieve the disentanglement of implicit and explicit representations. 
Meanwhile, two-step disentanglement methods based on Autoencoder~\cite{hadad2018two} or VAE~\cite{zheng2019disentangling} are also proposed. 
However, all of these methods need to leverage (partial) label information. 
Besides, VAE usually can not achieve high-quality generation in real-world scenarios
~\cite{ghosh2019variational}. 
Therefore, several studies begin to capture discrete and continuous factors of variation based on GAN. InfoGAN~\cite{chen2016infogan} reveals the disentanglement of latent code by maximizing the mutual information between the latent code and the generated data, but it is not specifically designed for clustering. ClusterGAN~\cite{mukherjee2019clustergan} integrated GAN with an encoder network for clustering by creating a non-smooth latent space. However, it ignores the real conditional distribution, which leads to generate trivial latent features and less diverse samples. Unlike conventional conditional GANs~\cite{mirza2014conditional}, our proposed method integrates the Autoencoder and GAN by constructing two cycle-consistencies, and separates the latent variables into two parts without any labels. 



\section{Conclusion}

In this work, we present \ProjectName, a new conditional generation framework that can generate diverse samples without labels and directly obtain the cluster assignments without clustering methods. Unlike most existing latent space clustering algorithms, our method does not build `clustering-friendly' latent space explicitly and does not need extra clustering operation. Therefore, our method avoids the difficulty of integrating latent feature construction and clustering. Furthermore, our method does not disentangle class relevant features from class non-relevant features. The disentanglement in our method is targeted to extract ``cluster information'' from data. Although our method does not depend on any explicit distance calculation in the latent space, the distance between data may be implicitly defined by the neural networks. 

The two cycle-consistencies ($\mathbf{x} \rightarrow (\mathbf{z}_c$, $\mathbf{z}_n ) \rightarrow \mathbf{x}$, ($\mathbf{z}_c$, $\mathbf{z}_n$) $\rightarrow \mathbf{x} \rightarrow$ ($\mathbf{z}_c$, $\mathbf{z}_n$) ) in \ProjectName~can help avoid the triviality of $\mathbf{z}_n$, and then avoid the generation of low diversity images in some degree. We have used the real images to train the encoder-generation pair ($\mathbf{x} \rightarrow (\mathbf{z}_c$, $\mathbf{z}_n ) \rightarrow \mathbf{x}$), which can help the encoder to estimate the real conditional distribution. However, due to the unsupervised fashion of clustering, the conditional distribution $Q(\mathbf{z}_c | \mathbf{x})$ specified by the generator of GAN may not match well with the true conditional distribution $P(\mathbf{z}_c|\mathbf{x})$ in real data, which is the case in both ClusterGAN and our \ProjectName. This may be another reason for the low diversity conditional generation~\cite{gong2019twin}. Improving GAN to create more diverse images is an important task for future work.

{\small
\bibliographystyle{ieee_fullname}
\bibliography{egbib}
}

\appendix
\input{appendix.tex}

\end{document}

%% file: appendix.tex
\newpage
\onecolumn

\section{Appendix}
\label{appendix}

\subsection{Training algorithm}

\begin{algorithm}\label{algo:1}
\DontPrintSemicolon
  
\KwInput{$\theta$, $\psi$, $\phi$ initial parameters of $G_{\theta}$, $D_{\psi}$ and $E_{\phi}$, the dimension of latent code $d_n$, the number of clusters K, the batch size B, the number of critic iterations per end-to-end iteration M, the regularization parameters $\beta_1$ - $\beta_3$}
\KwOutput{The parameters of $G_{\theta}$, $D_{\psi}$ and $E_{\phi}$}
\KwData{Training data set $\mathbf{x}$}

   \While{not converged}
   {
        \For{i=1, \ldots, M}    
        { 
   		    Sample $\mathbf{z}_n \sim P(\mathbf{z}_n)$ a batch of random noise \; 
   		    Sample $\mathbf{z}_c$ a batch of random one-hot vectors \;
   		    $\mathbf{z} \gets (\mathbf{z}_c, \mathbf{z}_n)$ \;
   		    $\mathbf{x}_g \gets G_{\theta}(\mathbf{z})$\;   
   		    Sample $\mathbf{x}_r \sim P_x$ a batch of the training dataset \; 
        	$\psi \gets \nabla_{\psi}( D_{\psi}(\mathbf{x}_r) - D_{\psi}(\mathbf{x}_g) )$
        }   
   
   		Sample $\mathbf{z}_n \sim P(\mathbf{z}_n)$ a batch of random noise \; 
   		Sample $\mathbf{z}_c$ a batch of random one-hot vectors \;
   		$\mathbf{z} \gets (\mathbf{z}_c, \mathbf{z}_n)$, $\mathbf{x}_g \gets G_{\theta}(\mathbf{z})$ \;
   		$(\hat{\mathbf{z}}_c, \hat{\mathbf{z}}_n) \gets E_{\phi}(\mathbf{x}_g)$ \; 
   		$(\mathbf{z}_c^r, \mathbf{z}_n^r) \gets E_{\phi}(\mathbf{x}_r)$ \;
   		$\mathbf{z}^r \gets (\mathbf{z}_c^r, \mathbf{z}_n^r)$, 
   		$\hat{\mathbf{x}}_r \gets G_{\theta}(\mathbf{z}^r)$ \; 
   		$\theta \gets \nabla_{\theta}( - D_{\psi}(G_{\theta}(z)) + |\mathbf{x}_r - \hat{\mathbf{x}}_r| + \beta_1 \operatorname{MMD}(\mathbf{z}_n^r, \mathbf{z}_n) + \beta_2 ||\mathbf{z}_n -\hat{\mathbf{z}}_n||_2^2 + \beta_3 \mathcal{H}(\mathbf{z}_c, \hat{\mathbf{z}}_c))$ \;
   		$\phi \gets \nabla_{\phi}(|\mathbf{x}_r - \hat{\mathbf{x}}_r| + \beta_1 \operatorname{MMD}(\mathbf{z}_n^r, \mathbf{z}_n) + \beta_2 ||\mathbf{z}_n -\hat{\mathbf{z}}_n||_2^2 + \beta_3 \mathcal{H}(\mathbf{z}_c, \hat{\mathbf{z}}_c))$ \;   		
   }
\caption{The training procedure of \ProjectName.}
\end{algorithm}

\subsection{Implementation details}

\begin{table*}[ht]
\centering
\caption{The dimensions of $\mathbf{z}_c$ and $\mathbf{z}_n$ in \ProjectName \, for different datasets. Note that the dimension of one-hot discrete latent variables $\mathbf{z}_c$ is equal to the number of clusters.}
\label{table:dimension}
\begin{tabular}{@{}lcccccccc@{}}
\hline
Dataset   & MNIST & Fashion-10 & YTF & Pendigits & 10x\_73k & COIL-100 & CIFAR-10\\ 
\hline
$\mathbf{z}_c$  & 10 & 10 & 41 & 10  & 8  & 100 & 10\\
$\mathbf{z}_n$  & 25 & 40 & 60 & 5  & 30 & 100 & 128\\
\hline
\end{tabular}
\end{table*}

\begin{table}[th]
  \caption{The structure summary of the generator (G), discriminator (D), and encoder (E) in \ProjectName \, for different datasets.}
  \label{table:structure}
  \centering
  \begin{tabular}{lccccc}
  \hline
  Dataset  & Layer Type & G-1/D-4/E-4 & G-2/D-3/E-3 &   G-3/D-2/E-2 & G-4/D-1/E-1\\
  \hline
  MNIST & Conv-Deconv & $4 \times 4 \times 64$ & $4 \times 4 \times 128$ & - & -\\
  Fashion-10 & Conv-Deconv & $4 \times 4   \times 64$ & $4 \times 4 \times 128$ & - & -\\
  YTF & Conv-Deconv & $5 \times 5 \times   32$ & $5 \times 5 \times 64$ & $5 \times 5 \times 128$ & $5 \times 5 \times 256$ \\
  Pendigits & MLP & 256 & 256 & - & - \\
  10x\_73k & MLP & 256 & 256 & - & - \\
  \hline
  \end{tabular}
\end{table}

\begin{table*}[ht]
\centering
\caption{The network structure of the generator (G), discriminator (D), and encoder (E) for dSprites experiments from~\cite{kim2018disentangling}. We set the dimensions of continuous and noise variables to 5 as infoGAN-CR~\cite{lin2020infogan}.}
\label{table:dSprites}
\begin{tabular}{@{}ll@{}}
\hline
Discriminator $D$ / Encoder $E$ & Generator $G$ \\ 
\hline
Input 64 $\times$ 64 binary image  & Input $\in \mathbb{R}^{10}$ \\
4 $\times$ 4 conv. 32 lReLU. stride 2 & FC. 128 ReLU. batchnorm \\
4 $\times$ 4 conv. 32 lReLU. stride 2. batchnorm & FC. 4 $\times$ 4 $\times$ 64 ReLU. batchnorm \\
4 $\times$ 4 conv. 64 lReLU. stride 2. batchnorm & 4 $\times$ 4 upconv. 64 lReLU. stride 2. batchnorm \\
4 $\times$ 4 conv. 64 lReLU. stride 2. batchnorm & 4 $\times$ 4 upconv. 32 lReLU. stride 2. batchnorm \\
FC. 128 lReLU. batchnorm (*) & 4 $\times$ 4 upconv. 32 lReLU. stride 2. batchnorm \\
From *: FC. 1 sigmoid. (output layer for $D$) & 4 $\times$ 4 upconv. 1 sigmoid. stride 2 \\ 
From *: FC. 128 lReLU. batchnorm. FC 10 for $E$ & \\
\hline
\end{tabular}
\end{table*}

\begin{table*}
  \caption{The network structure for CIFAR-10 dataset from~\url{https://github.com/google/compare_gan}. The ResBlock is the resample of the residual block with downsampling and upsampling. The input shape of images is $32 \times 32 \times 3$. The kernel size is described in the format [$filter\_h$, $filter\_w$, $stride$] and the output shape is described as $h \times w \times channels$. }
  \label{table:cifar10}
  \centering
\begin{tabular}{@{}llllll@{}}
\hline
\multicolumn{3}{c}{Discriminator $D$} & \multicolumn{3}{c}{Generator $G$} \\
\cline{1-6}
LAYER & KERNEL & OUTPUT & LAYER & KERNEL & OUTPUT \\
\hline
ResBlock & [3,3,1] & $16 \times 16 \times 128$ & $z$ & - & 128 \\
ResBlock & [3,3,1] & $8 \times 8 \times 128$ & Linear & - & $4 \times 4 \times 256$ \\
ResBlock & [3,3,1] & $8 \times 8 \times 128$ & ResBlock & [3,3,1] & $8 \times 8 \times 256$ \\
ResBlock & [3,3,1] & $8 \times 8 \times 128$ & ResBlock & [3,3,1] & $16 \times 16 \times 256$ \\
ReLU, Mean Pooling & - & 128 & ResBlock & [3,3,1] & $32 \times 32 \times 256$ \\
Linear & - & 1 & BN, ReLU & - & $32 \times 32 \times 256$ \\
 &  &  & Conv, Sigmoid & [3,3,1] & $32 \times 32 \times 3$ \\
\hline

\end{tabular}
\end{table*}

Table~\ref{table:dimension} summarizes the dimensions of latent variables for different datasets. And Table~\ref{table:structure} summarizes the network structures for different datasets. For the image datasets (MNIST, Fashion-MNIST, and YTF), we employ the similar $G_{\theta}$ and $D_{\psi}$ of DCGAN~\cite{radford2015unsupervised} with conv-deconv layers, batch normalization and leaky ReLU activations with a slope of 0.2. The $E_{\phi}$ uses the same architecture as $D_{\psi}$ except for the last layer. For the Pendigits and 10x\_73k datasets, the $G_{\theta}$, $D_{\psi}$, and $E_{\phi}$ are the MLP with 2 hidden layers of 256 hidden units each. The model parameters have been initialized following the random normal distribution.

We evaluate our method on dSprites for disentanglement using the architectures shown in Table~\ref{table:dSprites}. The generator's Adam learning rate is set to 0.001 and The learning rates of discriminator and encoder are set to 0.002. The total number of epoches is 28.

We also evaluate our method on CIFAR-10 dataset in a fully unsupervised settings. The architectures are shown in Table~\ref{table:cifar10}. We use the Adam optimizer with a learning rate 0.0002 for the generator, the discriminator and the encoder ($\beta_1 = 0.5$, $\beta_2 = 0.999$). We train the model with 5 discriminator steps before each generator and encoder step. The dimension of $\mathbf{z}_n$ is fixed to 128, and the batch size is set to 64. The spectral normalization is used on both generator and discriminator. We use the same class-conditional BatchNorm in the generator as Lucic~\etal~\cite{lucic2019high}, to incorporate the category information from $\mathbf{z}_n$. For the encoder, we combine the pre-trained SimCLR~\cite{chen2020simple} model and trainable 2-layer MLP with hidden size 512 and output size 138 ( dimensions of $\mathbf{z}_n$ and $\mathbf{z}_c$). The self-supervised SimCLR model is pre-trained by following the official implementation~\footnote{\url{https://github.com/google-research/simclr}}. The reasons of choosing pre-trained SimCLR model are based on reducing the parameters of encoder, and improving training efficiency. Different from previous experiments, we apply the following regularization parameters on CIFAR-10 dataset: $\beta_1=\beta_2=1$, $\beta_3=1$.





\newpage




%% file: egpaper_for_review.bbl
\begin{thebibliography}{10}\itemsep=-1pt

\bibitem{alimoglu1996methods}
Fevzi Alimoglu and Ethem Alpaydin.
\newblock Methods of combining multiple classifiers based on different
  representations for pen-based handwritten digit recognition.
\newblock In {\em Proceedings of the Fifth Turkish Artificial Intelligence and
  Artificial Neural Networks Symposium (TAINN 96)}. Citeseer, 1996.

\bibitem{arbel2018gradient}
Michael Arbel, Dougal Sutherland, Miko{\l}aj Bi{\'n}kowski, and Arthur Gretton.
\newblock On gradient regularizers for mmd gans.
\newblock In {\em Advances in Neural Information Processing Systems}, pages
  6700--6710, 2018.

\bibitem{bengio2013representation}
Yoshua Bengio, Aaron Courville, and Pascal Vincent.
\newblock Representation learning: A review and new perspectives.
\newblock {\em IEEE transactions on pattern analysis and machine intelligence},
  35(8):1798--1828, 2013.

\bibitem{caron2018deep}
Mathilde Caron, Piotr Bojanowski, Armand Joulin, and Matthijs Douze.
\newblock Deep clustering for unsupervised learning of visual features.
\newblock In {\em Proceedings of the European Conference on Computer Vision
  (ECCV)}, pages 132--149, 2018.

\bibitem{chang2017deep}
Jianlong Chang, Lingfeng Wang, Gaofeng Meng, Shiming Xiang, and Chunhong Pan.
\newblock Deep adaptive image clustering.
\newblock In {\em Proceedings of the IEEE International Conference on Computer
  Vision}, pages 5879--5887, 2017.

\bibitem{chen2020simple}
Ting Chen, Simon Kornblith, Mohammad Norouzi, and Geoffrey Hinton.
\newblock A simple framework for contrastive learning of visual
  representations.
\newblock {\em arXiv preprint arXiv:2002.05709}, 2020.

\bibitem{chen2016infogan}
Xi Chen, Yan Duan, Rein Houthooft, John Schulman, Ilya Sutskever, and Pieter
  Abbeel.
\newblock Infogan: Interpretable representation learning by information
  maximizing generative adversarial nets.
\newblock In {\em Advances in neural information processing systems}, pages
  2172--2180, 2016.

\bibitem{chuang2006fuzzy}
Keh-Shih Chuang, Hong-Long Tzeng, Sharon Chen, Jay Wu, and Tzong-Jer Chen.
\newblock Fuzzy c-means clustering with spatial information for image
  segmentation.
\newblock {\em computerized medical imaging and graphics}, 30(1):9--15, 2006.

\bibitem{dilokthanakul2016deep}
Nat Dilokthanakul, Pedro~AM Mediano, Marta Garnelo, Matthew~CH Lee, Hugh
  Salimbeni, Kai Arulkumaran, and Murray Shanahan.
\newblock Deep unsupervised clustering with gaussian mixture variational
  autoencoders.
\newblock {\em arXiv preprint arXiv:1611.02648}, 2016.

\bibitem{ghasedi2017deep}
Kamran Ghasedi~Dizaji, Amirhossein Herandi, Cheng Deng, Weidong Cai, and Heng
  Huang.
\newblock Deep clustering via joint convolutional autoencoder embedding and
  relative entropy minimization.
\newblock In {\em Proceedings of the IEEE International Conference on Computer
  Vision}, pages 5736--5745, 2017.

\bibitem{ghosh2019variational}
Partha Ghosh, Mehdi~SM Sajjadi, Antonio Vergari, Michael Black, and Bernhard
  Sch{\"o}lkopf.
\newblock From variational to deterministic autoencoders.
\newblock {\em arXiv preprint arXiv:1903.12436}, 2019.

\bibitem{gong2019twin}
Mingming Gong, Yanwu Xu, Chunyuan Li, Kun Zhang, and Kayhan Batmanghelich.
\newblock Twin auxilary classifiers gan.
\newblock In {\em Advances in Neural Information Processing Systems}, pages
  1328--1337, 2019.

\bibitem{goodfellow2014generative}
Ian Goodfellow, Jean Pouget-Abadie, Mehdi Mirza, Bing Xu, David Warde-Farley,
  Sherjil Ozair, Aaron Courville, and Yoshua Bengio.
\newblock Generative adversarial nets.
\newblock In {\em Advances in neural information processing systems}, pages
  2672--2680, 2014.

\bibitem{gretton2012kernel}
Arthur Gretton, Karsten~M Borgwardt, Malte~J Rasch, Bernhard Sch{\"o}lkopf, and
  Alexander Smola.
\newblock A kernel two-sample test.
\newblock {\em Journal of Machine Learning Research}, 13(Mar):723--773, 2012.

\bibitem{gulrajani2017improved}
Ishaan Gulrajani, Faruk Ahmed, Martin Arjovsky, Vincent Dumoulin, and Aaron~C
  Courville.
\newblock Improved training of wasserstein gans.
\newblock In {\em Advances in neural information processing systems}, pages
  5767--5777, 2017.

\bibitem{guo2017improved}
Xifeng Guo, Long Gao, Xinwang Liu, and Jianping Yin.
\newblock Improved deep embedded clustering with local structure preservation.
\newblock In {\em IJCAI}, pages 1753--1759, 2017.

\bibitem{hadad2018two}
Naama Hadad, Lior Wolf, and Moni Shahar.
\newblock A two-step disentanglement method.
\newblock In {\em Proceedings of the IEEE Conference on Computer Vision and
  Pattern Recognition}, pages 772--780, 2018.

\bibitem{haeusser2018associative}
Philip Haeusser, Johannes Plapp, Vladimir Golkov, Elie Aljalbout, and Daniel
  Cremers.
\newblock Associative deep clustering: Training a classification network with
  no labels.
\newblock In {\em German Conference on Pattern Recognition}, pages 18--32.
  Springer, 2018.

\bibitem{heusel2017gans}
Martin Heusel, Hubert Ramsauer, Thomas Unterthiner, Bernhard Nessler, and Sepp
  Hochreiter.
\newblock Gans trained by a two time-scale update rule converge to a local nash
  equilibrium.
\newblock In {\em Advances in neural information processing systems}, pages
  6626--6637, 2017.

\bibitem{ji2019invariant}
Xu Ji, Jo{\~a}o~F Henriques, and Andrea Vedaldi.
\newblock Invariant information clustering for unsupervised image
  classification and segmentation.
\newblock In {\em Proceedings of the IEEE International Conference on Computer
  Vision}, pages 9865--9874, 2019.

\bibitem{Karras2020ada}
Tero Karras, Miika Aittala, Janne Hellsten, Samuli Laine, Jaakko Lehtinen, and
  Timo Aila.
\newblock Training generative adversarial networks with limited data.
\newblock In {\em Proc. NeurIPS}, 2020.

\bibitem{Karras2019stylegan2}
Tero Karras, Samuli Laine, Miika Aittala, Janne Hellsten, Jaakko Lehtinen, and
  Timo Aila.
\newblock Analyzing and improving the image quality of {StyleGAN}.
\newblock In {\em Proc. CVPR}, 2020.

\bibitem{kim2018disentangling}
Hyunjik Kim and Andriy Mnih.
\newblock Disentangling by factorising.
\newblock {\em arXiv preprint arXiv:1802.05983}, 2018.

\bibitem{krizhevsky2009learning}
Alex Krizhevsky, Geoffrey Hinton, et~al.
\newblock Learning multiple layers of features from tiny images.
\newblock 2009.

\bibitem{lecun1998gradient}
Yann LeCun, L{\'e}on Bottou, Yoshua Bengio, Patrick Haffner, et~al.
\newblock Gradient-based learning applied to document recognition.
\newblock {\em Proceedings of the IEEE}, 86(11):2278--2324, 1998.

\bibitem{lee1999learning}
Daniel~D Lee and H~Sebastian Seung.
\newblock Learning the parts of objects by non-negative matrix factorization.
\newblock {\em Nature}, 401(6755):788, 1999.

\bibitem{li2015generative}
Yujia Li, Kevin Swersky, and Rich Zemel.
\newblock Generative moment matching networks.
\newblock In {\em International Conference on Machine Learning}, pages
  1718--1727, 2015.

\bibitem{lin2020infogan}
Zinan Lin, Kiran Thekumparampil, Giulia Fanti, and Sewoong Oh.
\newblock Infogan-cr and modelcentrality: Self-supervised model training and
  selection for disentangling gans.
\newblock In {\em International Conference on Machine Learning}, pages
  6127--6139. PMLR, 2020.

\bibitem{lucic2019high}
Mario Lucic, Michael Tschannen, Marvin Ritter, Xiaohua Zhai, Olivier Bachem,
  and Sylvain Gelly.
\newblock High-fidelity image generation with fewer labels.
\newblock {\em arXiv preprint arXiv:1903.02271}, 2019.

\bibitem{maaten2008visualizing}
Laurens van~der Maaten and Geoffrey Hinton.
\newblock Visualizing data using t-sne.
\newblock {\em Journal of machine learning research}, 9(Nov):2579--2605, 2008.

\bibitem{macqueen1967some}
James MacQueen et~al.
\newblock Some methods for classification and analysis of multivariate
  observations.
\newblock In {\em Proceedings of the fifth Berkeley symposium on mathematical
  statistics and probability}, volume~1, pages 281--297. Oakland, CA, USA,
  1967.

\bibitem{makhzani2015adversarial}
Alireza Makhzani, Jonathon Shlens, Navdeep Jaitly, Ian Goodfellow, and Brendan
  Frey.
\newblock Adversarial autoencoders.
\newblock {\em arXiv preprint arXiv:1511.05644}, 2015.

\bibitem{mao2019mode}
Qi Mao, Hsin-Ying Lee, Hung-Yu Tseng, Siwei Ma, and Ming-Hsuan Yang.
\newblock Mode seeking generative adversarial networks for diverse image
  synthesis.
\newblock In {\em Proceedings of the IEEE Conference on Computer Vision and
  Pattern Recognition}, pages 1429--1437, 2019.

\bibitem{mathieu2016disentangling}
Michael~F Mathieu, Junbo~Jake Zhao, Junbo Zhao, Aditya Ramesh, Pablo
  Sprechmann, and Yann LeCun.
\newblock Disentangling factors of variation in deep representation using
  adversarial training.
\newblock In {\em Advances in Neural Information Processing Systems}, pages
  5040--5048, 2016.

\bibitem{matthey2017dsprites}
Loic Matthey, Irina Higgins, Demis Hassabis, and Alexander Lerchner.
\newblock dsprites: Disentanglement testing sprites dataset, 2017.

\bibitem{mirza2014conditional}
Mehdi Mirza and Simon Osindero.
\newblock Conditional generative adversarial nets.
\newblock {\em arXiv preprint arXiv:1411.1784}, 2014.

\bibitem{mukherjee2019clustergan}
Sudipto Mukherjee, Himanshu Asnani, Eugene Lin, and Sreeram Kannan.
\newblock Clustergan: Latent space clustering in generative adversarial
  networks.
\newblock In {\em Proceedings of the AAAI Conference on Artificial
  Intelligence}, volume~33, pages 4610--4617, 2019.

\bibitem{nene1996columbia}
Sameer~A Nene, Shree~K Nayar, Hiroshi Murase, et~al.
\newblock Columbia object image library (coil-20).
\newblock 1996.

\bibitem{niu2020gatcluster}
Chuang Niu, Jun Zhang, Ge Wang, and Jimin Liang.
\newblock Gatcluster: Self-supervised gaussian-attention network for image
  clustering.
\newblock In {\em European Conference on Computer Vision}, pages 735--751.
  Springer, 2020.

\bibitem{odena2017conditional}
Augustus Odena, Christopher Olah, and Jonathon Shlens.
\newblock Conditional image synthesis with auxiliary classifier gans.
\newblock In {\em International conference on machine learning}, pages
  2642--2651. PMLR, 2017.

\bibitem{patacchiola2019autoencoders}
Massimiliano Patacchiola, Patrick Fox-Roberts, and Edward Rosten.
\newblock Y-autoencoders: disentangling latent representations via
  sequential-encoding.
\newblock {\em arXiv preprint arXiv:1907.10949}, 2019.

\bibitem{radford2015unsupervised}
Alec Radford, Luke Metz, and Soumith Chintala.
\newblock Unsupervised representation learning with deep convolutional
  generative adversarial networks.
\newblock {\em arXiv preprint arXiv:1511.06434}, 2015.

\bibitem{shaham2018spectralnet}
Uri Shaham, Kelly Stanton, Henry Li, Boaz Nadler, Ronen Basri, and Yuval
  Kluger.
\newblock Spectralnet: Spectral clustering using deep neural networks.
\newblock {\em arXiv preprint arXiv:1801.01587}, 2018.

\bibitem{tolstikhin2017wasserstein}
Ilya Tolstikhin, Olivier Bousquet, Sylvain Gelly, and Bernhard Schoelkopf.
\newblock Wasserstein auto-encoders.
\newblock {\em arXiv preprint arXiv:1711.01558}, 2017.

\bibitem{tschannen2018recent}
Michael Tschannen, Olivier Bachem, and Mario Lucic.
\newblock Recent advances in autoencoder-based representation learning.
\newblock {\em arXiv preprint arXiv:1812.05069}, 2018.

\bibitem{wang2019dominant}
Chu Wang, Marcello Pelillo, and Kaleem Siddiqi.
\newblock Dominant set clustering and pooling for multi-view 3d object
  recognition.
\newblock {\em arXiv preprint arXiv:1906.01592}, 2019.

\bibitem{wang2009mean}
Zhou Wang and Alan~C Bovik.
\newblock Mean squared error: Love it or leave it? a new look at signal
  fidelity measures.
\newblock {\em IEEE signal processing magazine}, 26(1):98--117, 2009.

\bibitem{wang2004image}
Zhou Wang, Alan~C Bovik, Hamid~R Sheikh, and Eero~P Simoncelli.
\newblock Image quality assessment: from error visibility to structural
  similarity.
\newblock {\em IEEE transactions on image processing}, 13(4):600--612, 2004.

\bibitem{wolf2011face}
Lior Wolf, Tal Hassner, and Itay Maoz.
\newblock Face recognition in unconstrained videos with matched background
  similarity.
\newblock In {\em Proceedings of the IEEE Conference on Computer Vision and
  Pattern Recognition}, pages 529--534, 2011.

\bibitem{xiao2017fashion}
Han Xiao, Kashif Rasul, and Roland Vollgraf.
\newblock Fashion-mnist: a novel image dataset for benchmarking machine
  learning algorithms.
\newblock {\em arXiv preprint arXiv:1708.07747}, 2017.

\bibitem{xie2016unsupervised}
Junyuan Xie, Ross Girshick, and Ali Farhadi.
\newblock Unsupervised deep embedding for clustering analysis.
\newblock In {\em International conference on machine learning}, pages
  478--487, 2016.

\bibitem{yang2017towards}
Bo Yang, Xiao Fu, Nicholas~D Sidiropoulos, and Mingyi Hong.
\newblock Towards k-means-friendly spaces: Simultaneous deep learning and
  clustering.
\newblock In {\em Proceedings of the 34th International Conference on Machine
  Learning}, volume~70, pages 3861--3870. JMLR.org, 2017.

\bibitem{yang2016joint}
Jianwei Yang, Devi Parikh, and Dhruv Batra.
\newblock Joint unsupervised learning of deep representations and image
  clusters.
\newblock In {\em Proceedings of the IEEE Conference on Computer Vision and
  Pattern Recognition}, pages 5147--5156, 2016.

\bibitem{yang2019deep}
Xu Yang, Cheng Deng, Feng Zheng, Junchi Yan, and Wei Liu.
\newblock Deep spectral clustering using dual autoencoder network.
\newblock In {\em Proceedings of the IEEE Conference on Computer Vision and
  Pattern Recognition}, pages 4066--4075, 2019.

\bibitem{zheng2017massively}
Grace~XY Zheng, Jessica~M Terry, Phillip Belgrader, Paul Ryvkin, Zachary~W
  Bent, Ryan Wilson, Solongo~B Ziraldo, Tobias~D Wheeler, Geoff~P McDermott,
  Junjie Zhu, et~al.
\newblock Massively parallel digital transcriptional profiling of single cells.
\newblock {\em Nature communications}, 8:14049, 2017.

\bibitem{zheng2019disentangling}
Zhilin Zheng and Li Sun.
\newblock Disentangling latent space for vae by label relevant/irrelevant
  dimensions.
\newblock In {\em Proceedings of the IEEE Conference on Computer Vision and
  Pattern Recognition}, pages 12192--12201, 2019.

\end{thebibliography}
